# Large Language Model Enhanced Clustering for News Event Detection

Adane Nega Tarekegn

Department of Information Science and Media Studies, University of Bergen, Norway
(email: adane.tarekegn@uib.no)

*Abstract*— The news landscape is continuously evolving, with an ever-increasing volume of information from around the world. Automated event detection within this vast data repository is essential for monitoring, identifying, and analyzing significant news occurrences across diverse platforms. This paper presents an event detection framework that leverages Large Language Models (LLMs) combined with clustering analysis to detect news events from the Global Database of Events, Language, and Tone (GDELT). The framework enhances event clustering through both pre-event detection tasks (keyword extraction and text embedding) and post-event detection tasks (event summarization and topic labelling). We also evaluate the impact of various textual embeddings on the quality of clustering outcomes, ensuring robust news categorization. Additionally, we introduce a novel Cluster Stability Assessment Index (CSAI) to assess the validity and robustness of clustering results. CSAI utilizes multiple feature vectors to provide a new way of measuring clustering quality. Our experiments indicate that the use of LLM embedding in the event detection framework has significantly improved the results, demonstrating greater robustness in terms of CSAI scores. Moreover, post-event detection tasks generate meaningful insights, facilitating effective interpretation of event clustering results. Overall, our experimental results indicate that the proposed framework offers valuable insights and could enhance the accuracy of news analysis and reporting.

## I. Introduction

News events are everywhere and constantly changing, significantly shaping social, political, economic, and various other aspects of public life [1]. Over the past few years, the rapid evolution of the internet has profoundly changed how news is disseminated globally, leading to the generation of vast amounts of news data daily. The news media often highlights news events that capture public interest and create ongoing streams of news reports, offering diverse viewpoints on global issues. The process of detecting such events from a huge amount of web data involves identifying and organizing related sets of significant events that occur in the real world [2][3]. Organizing news data based on events can enhance the structuring and classification of news from various online sources, thereby improving the online experience for users [4]. In this study, we propose a clustering-based framework for global news event detection from the GDELT(Global Database of Events, Language, and Tone) [5] project news database by leveraging the recent advances in Large Language Models (LLMs) alongside a novel cluster validation approach. In recent years, LLMs have been used as a new method for improving and guiding the clustering process by identifying subtle semantic representations [6][7]. To automatically identify and generate more cohesive event clusters, we chose to employ LLMs at various stages of the event detection framework: pre-event detection (keyword extraction, representation of those keywords as embeddings) and post-event detection (summarization and cluster labelling). First, the KeyBERT approach [8] was used to generate keywords, and then OpenAI's GPT model was applied to refine and optimize the keyword extraction process. For the text embedding task, we utilize the LLM embedding model (*text-embedding-ada-002*) [9], a part of the GPT-3 family from OpenAI, specifically tailored to create text embeddings. For a better understanding of the final clustering results and to characterize each event cluster, we applied a text summarization technique using *GPT-3.5-turbo-instruct*, which is an advanced LLM model designed for various tasks. In addition, each event cluster is semantically assigned to different news topics by using the International Press Telecommunications Council (IPTC) taxonomy [10].

Evaluation and understanding of the quality of clustering results are both important and challenging in the unsupervised domain [11][12]. In this paper, we used stability-based evaluation, which involves assessing the consistency of clustering derived from applying the same clustering algorithm to several independent and identically distributed samples [13]. This evaluation method involves dividing the data into training and testing data points, where the training sets are used for cluster construction, enabling the prediction of cluster memberships for the test data points. In this process, clustering stability is determined based on the distance between the features of the training and testing dataset within each cluster. In general, our framework can improve the accuracy and depth of news reporting while maintaining journalistic integrity and ethical standards.

The key contributions are summarized as follows:
- We have curated a large dataset involving main steps, such as news aggregation, cleaning, and preprocessing.
- We propose a framework for detecting events from GDELT using embeddings and clustering algorithms.
- We incorporate LLMs in the framework to enhance the clustering process through keyword extraction, text embedding, summarization, and labelling.
- We introduce a novel stability-based cluster validation index to validate the quality of clustering solutions using the similarity scores of input feature vectors.

## II. RELATED WORKS

Event detection is the main task in the process of event extraction and analysis [4]. The aim is to find documents that contain a particular event of interest from a large collection of texts or to obtain a set of clusters within a collection of texts, with each cluster comprising articles that discuss the same event [2][14]. The task of event detection can be traced back to 1998, when a collaborative effort was made to define the problem within the broader field of topic detection and tracking (TDT) [4]. Since then, a significant range of algorithms has been developed to address the problem across various domains, including social media [15], by leveraging the advancements in text mining and natural language processing. The frequent changes and constant flux of news data streams often lead researchers to favour unsupervised methods for event detection [16]. As a result, many recent news event detection tasks have focused on unsupervised learning [1] [2], [17], and [18], which employ clustering algorithms with traditional textual embedding.

In recent years, the emergence of LLMs has provided a new opportunity for improving tasks, such as data analysis, information retrieval, and question-answering systems. However, there is a paucity of adequate research on using LLMs to improve clustering and assess their effectiveness in event detection. In this paper, we aim to study the impact of LLMs on enhancing event clustering through both pre-event detection and post-event detection tasks. In the pre-event detection (or pre-clustering) task, we used LLMs for keyword generation and text embedding that identifies subtle semantic connections. In the post-detection task, LLMs were used for event labelling, summarization, and ITPC topic identification. Moreover, we evaluate the clustering results using a stability-based cluster validation index, a newly proposed method designed to assess both the robustness and validation of event clustering.

## III. STUDY FRAMEWORK

This section presents an overview of the overall architecture of the proposed framework for news event detection, as illustrated in Figure 1. The framework begins with the acquisition of the GDELT dataset and preprocessing, which involves tasks such as eliminating special characters, unnecessary words, symbols, and digits. Subsequently, keyword extraction and conceptual embedding of documents are executed using LLM and the traditional keyBERT model. UMAP representation of the dataset is performed to visualize the embedded feature vectors and mitigate processing time and storage complexity. The next phase of the pipeline involves selecting and applying a clustering algorithm to generate internally coherent clusters with distinct characteristics. The next important task is assessing the robustness and validation of the generated clusters, which is accomplished through a stability-based assessment of clustering results. Finally, we leverage LLM to interpret the clusters, which involves labelling, summarizing, and extracting the event groups with IPTC topics.

### A. Dataset and Preprocessing

A collection of news documents (about 15,000 news articles) was collected from a massive and regularly updated dataset of online, TV and news reporting from GDELT [5]. The GDELT project monitors global media through diverse perspectives, capturing and analyzing elements like themes, sentiments, geographic locations, and occurrences. It integrates translation across different languages and tracks emotions and themes in each news piece. GDELT is an example of a publicly accessible huge textual dataset which is available on a cloud platform. Within GDELT, the Global Knowledge Graph (GKG) serves as a main component, housing essential data such as sentiment scores, themes, and locations derived from newspaper articles. The GKG can be accessed through the use of Python and GDELT API, which analyze global newspaper articles in real time [19].

The news data collected from GDLET requires preprocessing before it can be analyzed to improve the efficiency and accuracy of learning models. In such cases, automating the preprocessing pipeline can help simplify the process [20]. We carried out essential data preprocessing steps, which included reformatting the dataset, normalizing it by converting it to lowercase and reducing noise by removing irrelevant information such as hyperlinks, symbols, and extraneous characters. Removing these elements is essential to prevent the feature space from becoming unnecessarily large, which could compromise computational efficiency and quality of results [21].

### B. LLMs in Clustering

In recent years, notable progress has been seen in LLMs, such as GPT-4 [22], and LLaMA-2 [23], which have shown remarkable abilities in zero-shot and few-shot learning. These models have different applications across diverse domains, effectively tackling tasks ranging from chatbots to language translation and content generation. In this work, we propose to use LLM for various tasks in the clustering process, including keyword extraction, text embedding, and interpretation of cluster contents by generating summaries and IPTC topic categories. Keyword extraction can be considered one of the main tasks in natural language processing, playing a pivotal role across diverse applications like content summarization, information retrieval, and clustering [24]. We used LLM to extract keywords that provide overarching insights, themes, or descriptions from the content and then applied various embeddings to each of these keywords. As a result, each news document is first processed by the keyword extraction model. We employ the *keyBERT* and *"GPT-3.5-turbo"* models for extracting and refining a list of keywords and phrases as well as to describe and label each cluster.

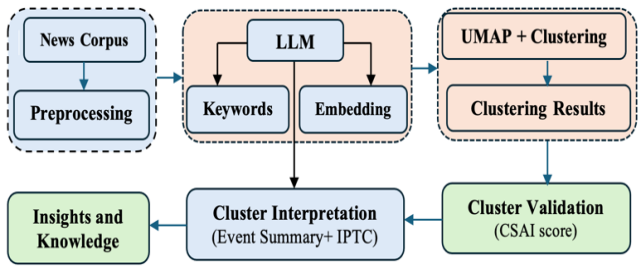

Fig.1. High-level overview of our framework for event detection

The generated keywords are then encoded by an embedding model. Embeddings can capture the semantic meaning and syntactic information in a text. Applying the right embeddings to represent textual information is essential to achieve optimal outcomes. In this study, OpenAI's state-of-the-art text embedding model *(text-embedding-ada-002)* [9] is used, which can generate high-dimensional vectors that effectively capture semantic similarities between words and phrases. According to recent studies on text clustering, OpenAI embeddings show better performance compared to other embeddings [25], and LLMs are generally good at improving cluster quality [6], despite their higher computational costs and complexity. Unlike shallow representations used in traditional methods, such as term frequency-inverse document frequency (TF-IDF), current embedding models leverage advanced techniques to embed text into continuous vector spaces where similar meanings are encoded closer together [26]. We have also explored other text embedding techniques, such as BERT embedding and Glove, and compared the results with those of LLM embedding. To characterize each of the event clusters, we applied text summarization and event categorization based on the IPTC taxonomy using the same family of GPT models (GPT-3.5-turbo-instruct). This can make the clusters understandable and aid in interpreting the characteristics and profiles of the clustered events.

*C. Dimensionality Reduction*

Dimensionality reduction aims to remove noise from data dimensions, enhancing clustering accuracy while also lowering computational costs. Furthermore, applying clustering to original dimensions can be difficult for higher-dimensional data. In this paper, we primarily used UMAP (Uniform Manifold Approximation and Projection) [27] to represent the reduced version of our data. Unlike other commonly used methods, such as PCA and t-SNE [28], UMAP preserves more global structures while also enhancing processing speed, making it a better choice for high-dimensional data visualization and analysis. UMAP has the capability to maintain the overall structure of the data (global structure) as well as the relationships between individual points (local structure) [29]. It is aimed at processing large amounts of information efficiently, providing faster computation times and making it suitable for real-world applications. One of the main parameters of UMAP includes the n-neighbours parameter, which determines the number of neighbouring points considered when constructing the local manifold.

On the other hand, the 'min-dist' parameter controls how tightly points are packed in the low-dimensional space, affecting clustering tightness. Lower values promote finer topological structures, while higher values result in looser point clustering, emphasizing the preservation of broader topological structures [27]. A 2D projection of the data using the t-SNE, and UMAP methods is shown in Figure 2.

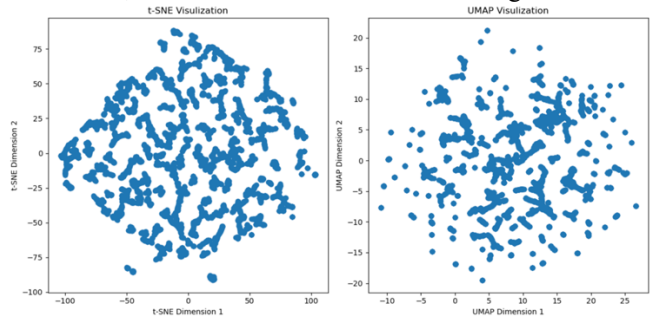

Fig. 2. 2D projections of the GDELT data via t-SNE (left), and UMAP (right)

*D. Clustering Algorithms*

To identify patterns in the vector representations of news text, we use various clustering algorithms and evaluate their performance. We use agglomerative hierarchical clustering, k-means, HDBSCAN (hierarchical density-based spatial clustering of applications with noise) and the Gaussian mixture model, which have different ways of grouping data into finite sets of categories [30][31]. Hierarchical clustering methods are categorized into agglomerative and divisive approaches. In this paper, agglomerative clustering is used, which works by assuming each data point is an individual cluster, and clusters are successively merged based on their similarity until a single cluster or a desired number of clusters is achieved [32]. Among partitioning-based clustering techniques, K-means is one of the most widely used. The reasons behind the popularity of K-means clustering are the ease of implementation, simplicity, efficiency, and empirical success [33]. However, it is affected by its initial configuration and can converge to local optima. HDBSCAN, on the other hand, can identify clusters with irregular shapes. It does this by modelling clusters as dense areas in the data space that are separated by sparser areas [34]. Instead of finding clusters with specific shapes, HDBSCAN identifies denser regions compared to their surroundings. Unlike K-means clustering, HDBSCAN does not need the user to select the number of clusters to be generated. The Gaussian mixture model is a model-based approach that seeks to optimize the fit between the data and a mathematical representation. It assumes that the data originates from a mixture of underlying probability distributions [35]. Moreover, these methods enable the automatic determination of the number of clusters using statistical methods, accounting for noise, thereby producing a robust clustering method.

IV. QUALITY ASSESSMENT AND EMBEDDINGS

This section presents the experimental results, evaluation of clustering algorithms, and impact of various embeddings on event clustering, along with a detailed discussion.

## A. Evaluation of Clustering Results

Cluster validation is a critical step in the unsupervised learning process, used to assess the validity and performance of clustering algorithms. Unlike supervised learning, where model evaluation metrics, such as accuracy, recall, and F-score [36][37] are well-defined, cluster validation presents unique challenges due to the absence of ground truth labels, subjectivity in evaluation, and the algorithm's sensitivity to initialization and small changes in the data. To address the challenges and assess the quality of generated event clusters, we proposed a New Cluster Stability Assessment Index (CSAI), which is based on the similarity of feature vectors between the training data used to create clusters and the validation data used to verify the stability of clusters. CSAI is based on the idea of prior work on multi-label data clustering [38]. However, instead of using a set of output labels and their associated probabilities to compute cluster indices, we rely on a set of input features and their similarities for inference on textual news datasets. Moreover, in CSAI, clusters are formed by partitioning the training part of the dataset into multiple subsamples, with no portion reserved for validation purposes. Evaluation of results is conducted separately using a dedicated validation dataset. CSAI employs normalized root mean squared error (NRMSE) as the distance measure between feature values in the validation dataset and feature scores in the training dataset corresponding to a specific cluster. Formally, CSAI is expressed in (1), while the dataset, algorithm, and source code for the proposed cluster validation index (CSAI) will be publicly available.

$$\text{CSAI} = \frac{1}{k}\sum_{j=1}^{K}\frac{1}{N}\sum_{n=1}^{N}\left(\frac{1}{(T_{max}-T_{min})}\sqrt{\frac{1}{F}\sum_{i=1}^{F}(V_i - T_i)^2}\right) \quad (1)$$

where K represents the total number of partitions in the training data, and N is the total number of clusters generated from each partition. $T_{max}$ and $T_{min}$ denoting the maximum and minimum values of features in the training data, respectively. F is the total number of features, $T_i$ and $V_i$ represent the numeric value of $i^{th}$ feature in the training and validation set, respectively. When the CSAI values are low, it indicates greater stability in the outputs of the clustering algorithm. High cluster stability occurs when minor variations in the dataset do not affect the cluster memberships.

CSAI addresses the limitations of existing clustering evaluation metrics by providing an assessment of both the quality and stability of clusters. This dual focus makes it a more reliable and robust metric, especially in real-world applications where data variations and the need for reproducibility are common challenges. Thus, CSAI can provide a better alterative over traditional metrics like the Calinski-Harabasz index.

## B. Effect of Embeddings on Clustering Results

We carried out a set of experiments to select the most effective embedding approach for our event detection framework. We use the commonly used embedding methods, such as TF-IDF [39], GloVe [40] and BERT [41] and compare them with the state-of-the-art LLM embedding. The TF-IDF representation emphasizes the significance of words within the dataset by assigning a vector to each word [39]. GloVe word embeddings utilize a co-occurrence matrix, where each element signifies the frequency of two words appearing together within a specified context window [40]. On the other hand, BERT embeddings leverage transformer-based bidirectional encoders to encode text into numerical representations, which have been commonly used in various tasks such as text clustering [25]. We evaluate these embedding methods against the LLM embedding model using different clustering algorithms. We also aim to identify which clustering algorithm performs best on which embedding type. For each embedding type, we apply five clustering algorithms: two from the partitioning group (k-means and k-medoid), one hierarchical (agglomerative clustering), one model-based (gaussian mixture model (GMM)), and one density-based (HDBSCAN). We then evaluate the performance of each algorithm using our proposed index (CSAI), as shown in Figure 3. We aim to investigate and select the best text embedding technique and clustering algorithm for our event detection framework.

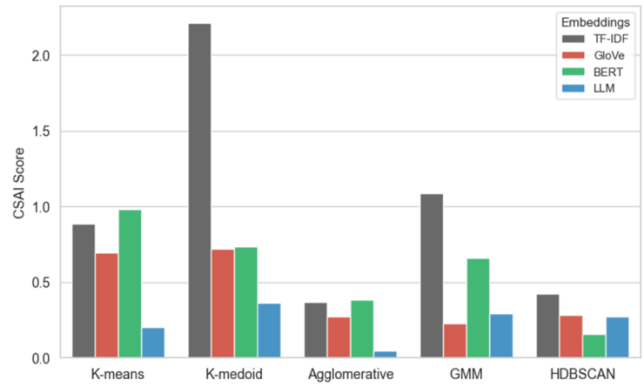

Fig. 3. Text clustering results based on CSAI scores on various word embedding models.

The top-performing algorithm was identified by selecting the one with the lowest CSAI score for each type of embedding. Our results indicate that the LLM embedding model has consistently achieved the lowest (i.e. the best) CSAI scores across all clustering algorithms, except for HDBSCAN, where it worked well with the BERT embedding. Among the clustering algorithms, agglomerative clustering combined with the LLM embeddings achieved the best CSAI scores, followed by k-means clustering on the same embeddings. Model-based clustering (GMM) has also shown better performance with the LLM embedding in terms of CSAI scores, while k-medoid has the worst CSAI score on TF-IDF and LLM embeddings. In terms of BERT, HDBSCAN has achieved better CSAI scores compared to other clustering models.

After identifying the best-performing clustering algorithm with the CSAI validity measure, the next step is to evaluate the robustness or stability of clustering results. Validity measures how accurately the clustering results capture the actual structure of the data, while robustness assesses the consistency of the clusters when subjected to different conditions like random initialization, resampling, perturbations, or variations in the algorithm [42]. Since the agglomerative hierarchical clustering algorithm showed the best results in terms of the CSAI validity analysis, we further examined the robustness of the clusters using a method known as stability analysis. We used the CSAI score again to assess the robustness of clusters across five different partitions of the data using various text embedding methods. The results in Figure 4 indicate that the agglomerative algorithm shows greater variability across the five partitions when using the BERT embedding, with the clustering model yielding a different CSAI value for each partition. There is also a slight variation in CSAI scores with GloVe, with the highest and lowest values observed on partitions 3 and 5, respectively. On the other hand, LLM and TF-IDF demonstrate greater stability across the five partitions. LLM delivers the best CSAI values for each partition, while TF-IDF has the worst CSAI scores for partitions 2 and 3. The average CSAI values for agglomerative hierarchical clustering across the different partitions were much better for LLM than other embeddings. Generally, when comparing embeddings based on CSAI scores, it is clear that clusters generated using the agglomerative algorithm with LLM embeddings demonstrate greater robustness compared to other embeddings across all five samples.

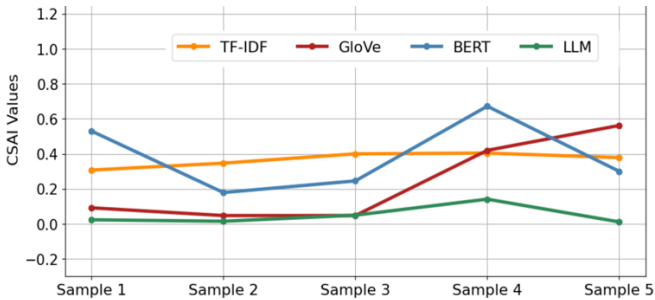

Fig. 4. Agglomerative clustering score in terms of CSAI across five samples using GloVe, BERT, and LLM embeddings.

V. CLUSTERING VISUALIZATION AND POST-DETECTION

A different number of event clusters have been discovered using various clustering algorithms on the GDELT news dataset. Determining the maximum number of clusters always varies from algorithm to algorithm. For example, in k-means clustering, we select the maximum number of clusters (k) using the elbow approach applied to the variance explained (i.e., the Within-Cluster-Sum of Squared Errors (WSS)) [43], where six potential clusters were identified by computing WSS against k, as depicted in Figure 5(a). Similarly, in hierarchical clustering, a dendrogram visually represents the arrangement of clusters, allowing us to find the optimal number of clusters by identifying effective separation points. HDBSCAN begins by creating a clustering hierarchy, which can be simplified into a tree showcasing the most significant clusters [44]. This hierarchy can easily be visualized as a traditional dendrogram or related representations from which only significant clusters are extracted. Figure 5 presents a scree plot with the elbow point for k-means clustering and a condensed cluster tree for HDBSCAN, along with their respective clustering plots.

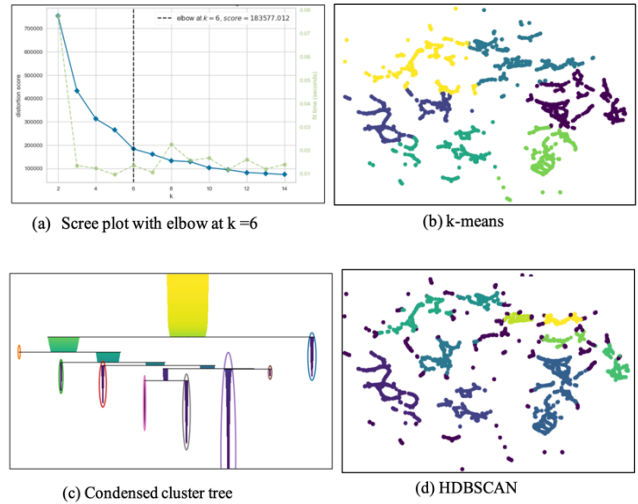

Fig. 5. Illustration of the optimal selection of clusters in k-means and HDBSCAN via the UMAP subspace and the corresponding clustering plots. Plot (a) shows a scree plot with 'elbow' to decide the best number of clusters (k) in k-means based on WSS, and (c) represents the condensed tree visualization help to indicate clusters identified by HDBSCAN. Plots (a) and (d) represent the 2D scatter plots of clustering results using k-means and HDBSCAN, respectively.

Besides visualization, post-event detection analysis is necessary for most applications, and it includes tasks such as event labelling, event summarization, and topic identification. Event clusters resulting from the clustering process must be labelled to facilitate the interpretation of the clustering outcomes. Interpretability is the ability of humans to understand the meanings of the outputs of the models in the context of domain knowledge [45]. Clustering in machine learning necessitates interpretability because the final outputs lack the ability to convey meaningful insights about cluster contents. In order to grasp the essence of the generated event clusters, this study explores the use of LLMs to label clusters using representative keywords, generate summaries for description, and categorize them according to the International Press Telecommunications Council (IPTC) taxonomy. Labelling entails assigning descriptive or explanatory terms to clusters, aiding in understanding their contents and characteristics. We also assigned the type of IPTC taxonomy to each cluster, facilitating the categorization of clusters according to predefined news topics. The other important post-detection task is event summarization, which can be essential for highlighting key aspects of an event. It can also support newsworthiness ranking and event validity identification, providing value to news users. Figure 6 displays a snapshot of the news event clusters alongside their corresponding keywords, ten-word summaries, and associated IPTC topics. The summary aids in conveying the results by providing a short highlight of each news event. As we can see from Figure 6, we can readily grasp the heterogeneity and characteristics of each cluster through concise summaries and IPTC groups. For example, clusters 0 and 1 have news stories related to business and the environment, respectively, while clusters 3 and 5 are focused on sports and technology news. Keywords and summaries provide quick insights into each article. For instance, cluster 1 includes an article discussing the Colorado River, revealing that 81% of its content is used by people for various activities. Such results suggested by our framework seem to be logical and provide additional assurance for the validity of clustering results.

| Cluster | Text | keywords | Summary | IPTC_Type |
|---|---|---|---|---|
| 3 | raphael lavoie s... | [ahl, lavoie, co... | Raphael Lavoie s... | Sports |
| 5 | the university o... | [oxford, univers... | Oxford Universit... | Technology/Science |
| 0 | secure energy se... | [analysts, ratin... | Secure Energy Se... | Business |
| 1 | for most of its ... | [freshwater, irr... | Study shows 81% ... | Environment |
| 4 | microsoft is cur... | [xbox, microsoft... | Microsoft workin... | Technology/Science |

Fig. 6. Event clusters with keywords, summary, and topic type

## VI. CONCLUSIONS

In this paper, we propose an event detection framework using the global GDLET news database. The framework comprises various components, including keyword extraction, text embedding, and event summarization aided by LLM, together with dimensionality reduction and clustering algorithms. We explore the use of different text embeddings to represent news articles and study how these embeddings enhance the effectiveness of the proposed framework. Our analysis compares the LLM embedding model with other techniques, such as TF-IDF, GloVe, and BERT, investigating their effectiveness for our event detection framework. Specifically, we evaluate how these embeddings improve the results of clustering algorithms in terms of stability-based cluster validation index (CSAI). CSAI is introduced in this study to measure the quality of clustering results, which operates by initially assigning new data points to existing clusters constructed from the training dataset. Then, it computes the similarity between feature vectors in training and testing samples allocated to the same cluster. From the experimental results, LLM embedding shows better performance in terms of CSAI score, although they have a disadvantage in terms of inference time. Additionally, LLM contributes to enhancing cluster interpretability through summary generation and topic assignment for each event cluster. In the future, we plan to explore state-of-the-art AI approaches for analyzing event progression and monitoring, specifically focusing on societal events of interest, including the ones defined in the CAMEO ontology [46]. Moreover, additional research will be performed to evaluate the effectiveness of CSAI applicability across several domains and datasets, such as images and sensors, to gain valuable insights into CSAI's consistency, robustness, and computational efficiency.